\documentclass[acmsmall]{acmart}
\makeatletter
\def\runningfoot{\def\@runningfoot{}}
\def\firstfoot{\def\@firstfoot{}}
\makeatother
\renewcommand\footnotetextcopyrightpermission[1]{} 
\usepackage{fancyhdr}
\AtBeginDocument{%
  \providecommand\BibTeX{{%
    \normalfont B\kern-0.5em{\scshape i\kern-0.25em b}\kern-0.8em\TeX}}}

\usepackage{amssymb}
\usepackage{multirow}
\usepackage{xcolor}
\usepackage{colortbl}  
\newcommand{\ie}{\textit{i}.\textit{e}.}
\newcommand{\eg}{\textit{e}.\textit{g}.}

\newcommand{\etal}{\textit{et}.\textit{al}.}

\usepackage{pdfrender}

\usepackage{bm}
\usepackage{amsmath}
\usepackage[caption=false,font=footnotesize,labelfont=sf,textfont=sf]{subfig}
\usepackage[switch]{lineno}
\usepackage{url}
\usepackage{graphicx}
\usepackage{booktabs}
\usepackage{arydshln}

\usepackage{amsmath}
\usepackage{algorithmic}
\usepackage{array}
\usepackage{textcomp}
\usepackage{stfloats}
\usepackage{url}
\usepackage{verbatim}
\usepackage{graphicx}
\usepackage{balance}
\usepackage{hyperref}
\usepackage{multirow}
\usepackage{multicol}
\usepackage{enumitem}
\usepackage{arydshln} 
\usepackage{colortbl} 
\usepackage{pifont}
\usepackage{makecell}

\hypersetup{colorlinks,linkcolor={blue},citecolor={green},urlcolor={magenta}}

\definecolor{mygreen}{rgb}{0.09, 0.45, 0.27}
\definecolor{myred}{rgb}{1.0, 0.0, 0.0}
\definecolor{myyellow}{rgb}{1.0, 0.75, 0.0}
\definecolor{myblue}{rgb}{0.3, 0.55, 0.9}
\definecolor{mygray-bg}{gray}{0.9}

\begin{document}

\title{Improving Reference-based Distinctive Image Captioning with Contrastive Rewards}

\author{Yangjun Mao}
\affiliation{%
  \institution{Zhejiang University}
  \city{Hangzhou}
  \country{China}}
\email{maoyj0119@zju.edu.cn}

\author{Jun Xiao}
\affiliation{%
  \institution{Zhejiang University}
  \city{Hangzhou}
  \country{China}}
\email{junx@zju.edu.cn}

\author{Dong Zhang}
\affiliation{%
  \institution{The Hong Kong University of Science and Technology}
  \city{Kowloon}
  \country{Hong Kong SAR}}
\email{dongz@ust.hk}

\author{Meng Cao}
\affiliation{%
  \institution{Peking University}
  \city{Shenzhen}
  \country{China}}
\email{mengcao@pku.edu.cn}

\author{Jian Shao}
\affiliation{%
  \institution{Zhejiang University}
  \city{Hangzhou}
  \country{China}}
\email{jshao@zju.edu.cn}

\author{Yueting Zhuang}
\affiliation{%
  \institution{Zhejiang University}
  \city{Hangzhou}
  \country{China}}
\email{yzhuang@zju.edu.cn}

\author{Long Chen}
\authornote{Long Chen is the corresponding author.}
\affiliation{%
  \institution{The Hong Kong University of Science and Technology}
  \city{Kowloon}
  \country{Hong Kong SAR}}
\email{longchen@ust.hk}


\renewcommand{\shortauthors}{Mao, et al.}

\begin{abstract}
Distinctive Image Captioning (DIC) --- generating distinctive captions that describe the unique details of a target image --- has received considerable attention over the last few years. A recent DIC method proposes to generate distinctive captions by comparing the target image with a set of semantic-similar reference images, \ie, reference-based DIC (Ref-DIC). It aims to force the generated captions to distinguish between the target image and the reference image. Unfortunately, reference images used by existing Ref-DIC works are easy to distinguish: \emph{these reference images only resemble the target image at scene-level and have few common objects, such that a Ref-DIC model can trivially generate distinctive captions even without considering the reference images.} For example, if the target image contains objects ``\texttt{towel}'' and ``\texttt{toilet}'' while all reference images are without them, then a simple caption ``\texttt{A bathroom with a towel and a toilet}'' is distinctive enough to tell apart target and reference images. 
To ensure Ref-DIC models really perceive the unique objects (or attributes) in target images, we first propose two new Ref-DIC benchmarks. Specifically, we design a two-stage matching mechanism, which strictly controls the similarity between the target and reference images at the object-/attribute- level (v.s. scene-level). Secondly, to generate distinctive captions, we develop a Transformer-based Ref-DIC baseline \emph{TransDIC}. It not only extracts visual features from the target image, but also encodes the differences between objects in the target and reference images. Taking one step further, we propose a stronger \textbf{TransDIC++}, which consists of an extra contrastive learning module to make full use of the reference images. This new module is model-agnostic, which can be easily incorporated into various Ref-DIC architectures. Finally, for more trustworthy benchmarking, we propose a new evaluation metric named \emph{DisCIDEr} for Ref-DIC, which evaluates both the accuracy and distinctiveness of the generated captions. Experimental results demonstrate that our \textbf{TransDIC++} can generate distinctive captions. Besides, it outperforms several state-of-the-art models on the two new benchmarks over different metrics. 
\end{abstract}





\maketitle

\section{Introduction}
Image captioning, \ie, generating natural language descriptions to summarize the salient contents of a target image, has drawn much attention from the multimedia community. It has great impacts on many downstream applications, such as helping the blind people and developing navigation systems. However, as revealed in~\cite{dai2017towards, dai2017contrastive}, conventional image captioning models tend to generate over-generic captions or even identical captions when input images are similar. Obviously, these generic captions neglect the unique details of the target image. Recent captioning works~\cite{dai2017contrastive, luo2018discriminability, liu2018show, wang2020compare} begin to make generated captions more distinctive and ask these captions to describe more unique details of each target image, called Distinctive Image Captioning (DIC). 

\begin{figure}[t]
\centering
\includegraphics[width=\linewidth]{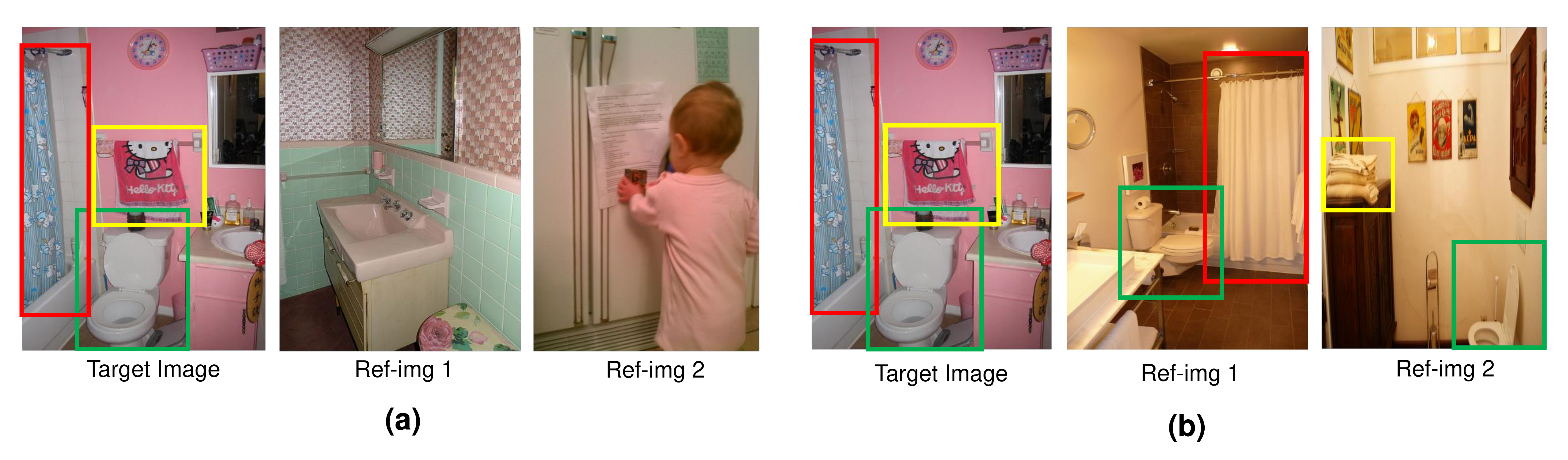}
\caption{(a): An example of constructed reference image group used in existing Ref-DIC work~\cite{wang2021group}. (b): Selected reference images for the same target image using our two-stage matching mechanism. We use the same colors to denote the same object categories in the different images (\eg, ``\texttt{towel}'' is with \textcolor{myyellow}{yellow} box, ``\texttt{shower curtain}'' is with \textcolor{myred}{red} box, and ``\texttt{toilet}'' is with \textcolor{mygreen}{green} box). We only show two reference images here. }
\label{fig_1_dataset_example}
\end{figure}

Currently, mainstream DIC works follow the same setting as plain image captioning: using one single image as input, and generating distinctive captions for each image, dubbed as Single-image DIC (\textbf{Single-DIC}). In this setting, they tend to generate a totally distinctive caption. By ``totally'', we mean the generated caption is asked to distinguish its corresponding image from all images in the dataset, \ie, dataset-level distinctiveness. To this end, they always resort to reinforcement learning and develop different distinctive rewards~\cite{luo2018discriminability,liu2018show}. However, this Single-DIC setting has two inherent issues: 1) It is difficult (or impossible) to generate a totally distinctive caption for the target image unless we describe all the details in the image. 2) Even for our humans, we still need some reference images when generating distinctive captions. For example in Fig.~\ref{fig_1_dataset_example}(b), without any reference images, people won't know what should be emphasized in the \emph{target image}, and may simply predict ``\texttt{A bathroom with a towel}'' for the image. In contrast,  they will focus on the unique colors of ``\texttt{towel}'' and ``\texttt{shower curtain}'', and predict ``\texttt{A bathroom with a pink towel and a blue shower curtain}'' when they use the ``\texttt{white shower curtain}'' in \emph{ref-img1} and ``\texttt{yellow towel}'' in \emph{ref-img2} as references.

For human-like distinctive captioning, recent work~\cite{wang2021group} proposes to study the DIC task based on a group of semantic-similar reference images, dubbed Reference-based DIC (\textbf{Ref-DIC}).
Different from Single-DIC, they use the target image and all reference images as input and these reference images will inform DIC models which parts of the target image should be emphasized. Compared to Single-DIC, the generated captions are only asked to distinguish the target image from the group of reference images, \ie, group-level distinctiveness. Unfortunately, the reference images used in existing Ref-DIC works~\cite{wang2021group} can be trivially distinguished: \emph{these reference images only resemble the target image at the scene-level and have few common objects, thus Ref-DIC models can simply generate distinctive captions even without considering the reference images.} For example in Fig.~\ref{fig_1_dataset_example}(a), \emph{target} and \emph{reference images} have no object in common (\eg, ``\texttt{towel}'', ``\texttt{shower curtain}'', or ``\texttt{toilet}''), each object in \emph{target image} is unique, such that the Ref-DIC model can trivially generate ``\texttt{a bathroom with a towel}'' to tell the target and reference images apart. 

\begin{figure}[t]
\centering
\includegraphics[width=\linewidth]{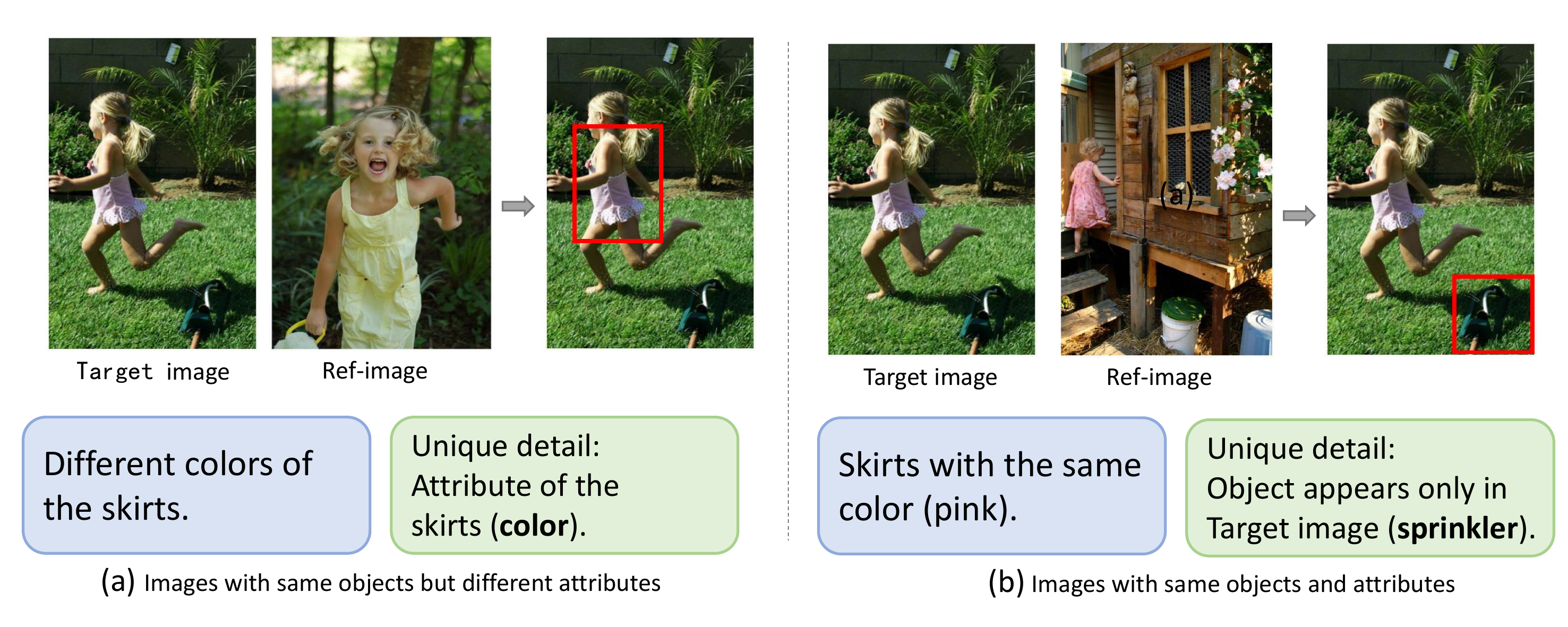}
\caption{(a): An example of target and reference images with common objects but different attributes. (b): An example of target and reference images with same objects and attributes.}
\label{fig_compare}
\end{figure}

To retrieve more reasonable reference images, we believe target and reference images should at least have some common objects. Therefore, we introduce two kinds of paradigms: 1) \emph{Images with the same objects but different attributes}. As shown in Fig.~\ref{fig_compare}(a), two girls are both wearing skirts but with different colors, the unique detail of the target image is the attribute of the skirts (color). 2) \emph{Images with identical objects}. In Fig.~\ref{fig_compare}(b), two girls have skirts with the same colors, so the unique detail of the target images is the sprinkler appears only in the target image. From the two above-mentioned examples, we show that unique details vary compared to different reference images. When using multiple reference images, we hope the model can focus on more specific visual details in the target image.

As mentioned above, reference images are crucial when defining the unique details in the target image. In this paper, we propose two new benchmarks for the Ref-DIC task: \textbf{COCO-DIC} and \textbf{Flickr30K-DIC}.
To strictly control the unique details between target and reference images, we propose a two-stage matching mechanism, which can measure image similarity at the object-/attribute- level (v.s. scene-level in~\cite{wang2021group}), and deliberately make target and reference images have some common objects. Under this mechanism, Ref-DIC models can learn to focus on the unique attributes and objects in the target image. As the example in Fig.~\ref{fig_1_dataset_example}(b), compared to \emph{ref-img1}, \emph{target image} has the unique attribute ``\texttt{blue}'' of ``\texttt{shower curtain}'' and the unique object ``\texttt{towel}''.

To achieve group-level distinctiveness, we propose to emphasize both unique attributes and objects in the target image. Thus, previously we propose a new Transformer-based captioning model named \textbf{TransDIC}, which directly gives each region in the target image (target regions) some region references when generating captions. Specifically,  we firstly find similar regions from reference images (reference regions) for each target region. Then, we send the target region and its corresponding reference regions into the \emph{Two-Flow} Module, which consists of a Target flow and a Target-Reference flow. The Target flow aims to encode target image features through self-attention blocks~\cite{vaswani2017attention}, while the Target-Reference flow enables cross-image interactions between target and reference images through a multi-layer co-attention. Different from the existing Ref-DIC work~\cite{wang2021group} which proposes an attention module to focus on unique objects in the target image, our TransDIC directly enables the feature interactions between target and reference images. 

On the other hand, distinctive reward~\cite{luo2018discriminability,liu2018show} has already been designed for the DIC task. Existing works mainly resort to an image-text retrieval model~\cite{faghri2017vse++}. Specifically, when a specific caption is generated, they send the caption along with its corresponding input image and a batch of semantically similar distractor images into the retrieval model. Based on the image-text similarity scores, the retrieval model informs the caption generator: the generated caption should have a higher score with the input image compared to distractors. These methods only take distractor images into consideration in the retrieval model, while the caption generator has never ``seen'' these distractors. We argue that there exists a large semantic gap between the retrieval model and the caption generator. Therefore, the calculated reward from the retrieval model may not fully inform the caption generator where to focus. Meanwhile, all these methods take the distractor images at image-level and fall to investigate the importance of each object.

To make full use of distractors (\ie, the reference images) and generate more distinctive captions, we design a new contrastive-learning-based (CL) module and a new reward DisReward for the Ref-DIC task. The whole model is dubbed as \textbf{TransDIC++}. Based on the reinforcement learning strategy, the reward encourages the model to generate better captions over negative samples using positive samples. Here, ``positive samples'' mean original target and reference images, and ``negative samples'' mean original target images and contaminated reference images. Different from the above-mentioned methods, our solution leverages both the target and reference images and the designed reward acts as a signal to inform the caption generator where to focus when generating captions. Furthermore, to take the investigation of the reference images at a fine-grained level, we design two kinds of masking strategies to construct negative samples: 1) Instance-level masking; 2) Image-level masking. They act as fine-grained and coarse-grained strategies that study the influence of reference images. Above all, the CL module is a plug-and-play module for the Ref-DIC task and can be easily applied to any multi-image input task.

Finally, to fully take advantage of ground-truth captions of reference images, we propose a new CIDEr-based~\cite{vedantam2015cider} metric termed as \textbf{DisCIDEr}. According to our definition of group-level distinctiveness, we believe frequently-used n-grams in ground-truth captions of reference images should be given less weight at evaluation time. The metric can not only directly evaluate the distinctiveness, but also preserve the accuracy advantage of CIDEr. Extensive experimental results on multiple Ref-DIC benchmarks (\eg, COCO-DIC and Flickr30K-DIC) have demonstrated the effectiveness of our proposed TransDIC and \textbf{TransDIC++}.

In summary, we make three contributions in this paper:

\begin{itemize}
    \item  We propose a complete pipeline for the Ref-DIC task, which consists of benchmarks for model training, specific model designs, and evaluation metrics.
    \item  We design a novel contrastive learning based module for the Ref-DIC task. The module is model agnostic, \ie, it is highly flexible and can be applied to other image captioning models. Meanwhile, we designed multiple ways to construct positive/negative samples to benefit the contrastive training paradigm.
    \item  We conduct extensive experiments on the new COCO-DIC and Flickr-DIC benchmarks. Extensive experimental results have shown the promising performance of our proposed methods.
\end{itemize}

\noindent \textbf{Highlights}. It is worth noting that this paper is a substantial extension of our previous conference publication on ACM Multimedia~\cite{mao2022rethinking}. Compared to the conference version, this manuscript has made three main improvements: 1) We explore a new contrastive learning based module and develop a new reward, DisReward, for Ref-DIC. The results show that the method further surpasses the original performance. 2) To further investigate the influence of reference images, we propose two kinds of masking strategy: instance-level and image-level masking, comprehensive experiments are conducted to demonstrate the effectiveness. 3) We apply our new module to other Ref-DIC models. More experimental results show our proposed CL module is promising and highly flexible to other Ref-DIC models.

\section{Related Work}
\subsection{Image Captioning}  Most modern image captioning models typically employ an encoder-decoder framework for caption generation~\cite{vinyals2015show,karpathy2015deep,johnson2016densecap,liu2021region,wang2021high}. Within this framework, many efforts have been made to improve the architecture, including attention mechanisms~\cite{xu2015show,chen2017sca,anderson2018bottom,huang2019attention,chen2021human}, graph convolution networks~\cite{yao2018exploring,yang2019auto,chen2020say}, and transformer-based models~\cite{li2019entangled,2020Meshed}. Meanwhile, another series of works explore different training objectives at the training stage. For example, Dai~\etal~\cite{dai2017towards} and Shetty~\etal~\cite{shetty2017speaking} leverage Generative Adversarial Network (GAN) to improve the diversity of generated captions. Some recent captioning works apply Reinforcement Learning (RL) to captioning and achieve great success~\cite{ranzato2015sequence,rennie2017self,liu2017improved,xu2019multi}. These models directly optimize  non-differentiable 
evaluation metrics (\eg, BLEU~\cite{papineni2002bleu}, CIDEr~\cite{vedantam2015cider}), which boost the caption generation procedure at the sentence-level.

\subsection{Distinctive Image Captioning (DIC)} Compared with conventional image captioning, DIC is a more challenging task, which tends to generate more informative and descriptive captions.
According to the stage they take effect, existing solutions can be coarsely divided into two categories: \emph{Inference-based} and \emph{Training-based} methods. Inference-based models mainly modify the caption decoding procedure at inference time and thus can be applied to any captioning architecture~\cite{vedantam2017context,wang2020towards}. In contrast, training-based methods, resort to different training objects~\cite{luo2018discriminability,liu2018show} or the progressive training procedure~\cite{liu2019generating}. Recently, some works begin to study the DIC task based on semantic-similar reference images. Specifically, Wang~\etal~\cite{wang2020compare} propose to assign higher weights to distinctive ground-truth captions at the training stage, Wang~\etal~\cite{wang2021group} use multiple images as input to emphasize distinctive objects. In this paper, we propose a co-attention based model to directly enable the feature interactions between target and reference images.

To evaluate the distinctiveness of generated captions in DIC, several new evaluation metrics are developed.  SPICE-U~\cite{wang2020towards} is designed for Single-DIC. CIDErBtw~\cite{wang2020compare} measures the distinctiveness of caption at the sentence-level similarity.  DisWordRate~\cite{wang2021group} directly evaluates the occurrences of distinctive words. In this paper, we develop a new metric named DisCIDEr for Ref-DIC. Compared to existing metrics, our metric fully explores the distinctiveness of each individual n-gram in ground-truth captions of the target image.

\subsection{Multi-input Image Captioning} Several captioning settings need multiple images as input. According to the number of input images, they can be divided into two categories: \emph{Two-image based} and \emph{Group-based} captioning.
Two-image based captioning tends to describe the common~\cite{suhr2018corpus} or different~\cite{tan2019expressing,park2019robust,yan2021l2c,qiu2021describing} parts between the two images. For example, the change captioning task takes before and after images as input and describes the changes between them.
Chen~\etal~\cite{chen2018groupcap} firstly model the relevance and diversity between target and reference images and aim to generate diverse captions for the target image. Li~\etal~\cite{li2020context} tend to describe a group of target images using another group of semantically similar images as references.

\subsection{Contrastive Learning}
Contrastive learning has been wildly applied in the field of representation learning~\cite{he2020momentum,chen2020simple,grill2020bootstrap,han2018face,chen2021exploring,zagoruyko2015learning}, they mainly resort to the architecture of Siamese Network and design different strategies to construct the negative samples. MoCo~\cite{he2020momentum} and SimCLR~\cite{chen2020simple} apply complex data augmentations on input images as negative samples. In contrast, BYOL\cite{grill2020bootstrap} and Simsiam\cite{chen2021exploring} explore an extra predictor head and stop-gradient strategy to relieve the explicit construction of negative samples. Inspired by their success, some image captioning works also utilize the contrastive training paradigm. The paradigm is also leveraged in the image captioning task. For example, Dai~\etal~\cite{dai2017contrastive} use mismatched image-caption pairs as negative samples and asks the model to tell apart them. However, these mismatched pairs are randomly generated, the selected captions for an image may be ``too mismatched'' and can be easily distinguished by the model. In this work, based on the Ref-DIC setting, we propose to mask specific proposals in reference images to form the negative samples.

\section{Ref-DIC Benchmarks}
In this section, we first formally define the Ref-DIC task. Then we describe our solution for Ref-DIC benchmarks construction. Finally, we provide details of our proposed COCO-DIC and Flickr30K-DIC benchmarks for Ref-DIC.

\subsection{Task Definition: Reference-based DIC}
Given a \textbf{target image} $I_t$ and a group of $K$ \textbf{reference images} $\mathcal{I}_r=\{I_i\}_{i=1}^K$ which are semantic-similar to $I_t$, Ref-DIC models aim to generate a natural language sentence $S=\{w_1,w_2,\ldots,w_T\}$. The generated sentence $S$ should not only correctly describe the target image $I_t$, but also contain sufficient details about $I_t$, so it can tell apart target and  reference images. For example in Fig.~\ref{fig_1_dataset_example}(b), given the target and reference images, Ref-DIC models aim to generate a distinctive caption ``\texttt{a bathroom with a pink towel, a blue shower curtain and a toilet}''. The detail ``\texttt{pink towel}'' is helpful to distinguish target image from \emph{ref-img2}~because the ``\texttt{towel}'' in \emph{ref-img2} is ``\texttt{white}''. On the contrary, predicting  ``\texttt{a bathroom with a towel and a toilet}'' fails to meet the requirements because it is suitable for both target and reference images.

\begin{figure*}[t]
\centering
\includegraphics[width=\linewidth]{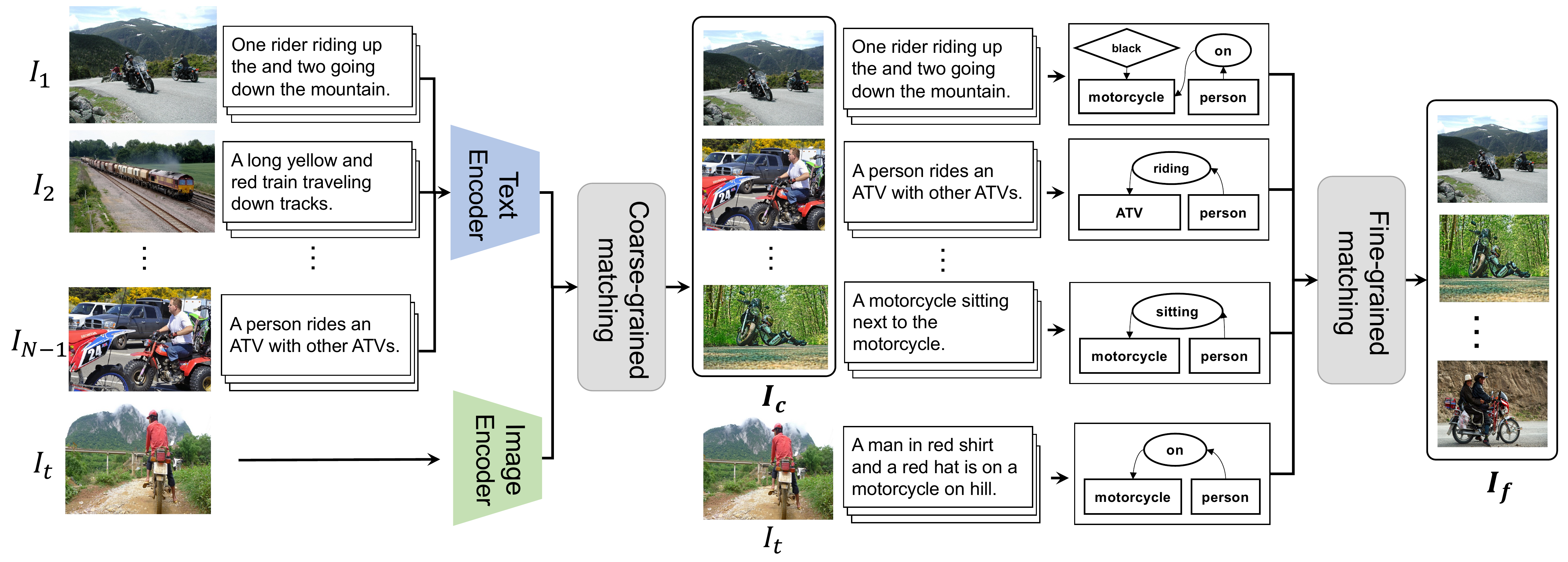}
\caption{The pipeline of our two-stage matching procedure. In the first stage, we calculate image-text similarity scores between target image $I_t$ and the captions of all other images in $\mathcal{D}$ through CLIP~\cite{radford2021learning} and construct a coarse-grained group $\mathcal{I}_c$ for target image $I_t$. In the second stage, we leverage scene graphs to calculate the object and attribute overlaps between images in $\mathcal{I}_c$ and $I_t$. We rearrange $\mathcal{I}_c$ according to their similarity scores with $I_t$, and finally get the fine-grained reference image group $\mathcal{I}_f$.}
\label{fig_2_group_construction}
\end{figure*}

\subsection{Ref-DIC Benchmarks Construction} \label{sec:sim_group}
Given a conventional image captioning dataset $\mathcal{D}$, suppose it contains $N$ images and each one has $M$ corresponding ground-truth captions. We build new Ref-DIC benchmarks based on $D$, by coupling each image (target image) with several semantic-similar reference images.
Specifically, each image in $\mathcal{D}$ will be regarded as a target image $I_t$, and all remaining $N-1$ images are termed as its \textbf{candidate reference images}. For each target image $I_t$, our goal is to retrieve $K$ reference images from its candidate reference images to construct the reference image group $\mathcal{I}_r$.

To achieve group-level distinctiveness, $I_t$ and retrieved $\mathcal{I}_r$ should have some common objects, such that $\mathcal{I}_r$ will inform the model to focus on the unique details in $I_t$. To this end, we design a two-stage matching mechanism. In the first stage, we construct a \textbf{coarse-grained group} $\mathcal{I}_c$ based on the image-text similarity score for each target image. Then in the second stage, we investigate fine-grained details of $I_t$ and $\mathcal{I}_c$, and construct a \textbf{fine-grained group} $\mathcal{I}_f$ based on $\mathcal{I}_c$. Finally, we select $K$ images out of $\mathcal{I}_f$ to construct the $\mathcal{I}_r$. We detailed introduce our two-stage matching pipeline below.

\subsubsection{Coarse-grained Group Construction} \label{sec:candidate group}
Following~\cite{wang2021group}, we use an image-text retrieval model to calculate similarity scores between images and texts. Specifically, as illustrated in the left side of Fig.~\ref{fig_2_group_construction}, we use a pre-trained  CLIP~\cite{radford2021learning} to firstly extract the visual feature of target image $I_t$ and text features of ground-truth captions from candidate reference images. Then, we perform the cosine similarity between the visual feature and all text features to get $(N-1)\times M$ scores. Finally, we select $||\mathcal{I}_c||$ captions with the highest scores, and their corresponding images are used as the coarse group $\mathcal{I}_c$ for $I_t$.
The CLIP-based matching mechanism can effectively filter out some obviously unrelated candidate images. However, since it encodes texts at the sentence level, several fine-grained details may be neglected when computing similarity. For example in Fig.~\ref{fig_2_group_construction}, ground-truth captions of candidate image $I_1$ contain the  ``\texttt{train}'', thus $I_1$ will be removed due to a low scene-level similarity score to $I_t$. Meanwhile, $I_{N-1}$ is considered similar to $I_t$ by the CLIP because both of them describe the scene of ``\texttt{someone is riding a vehicle}''. However, they resemble each other only at the sense-level and do not contain any common objects (\eg, ``\texttt{motorcycle}'' in $I_{t}$ and ``\texttt{ATV}'' in $I_{N-1}$ are totally different objects).

\begin{figure}[t]
\centering
\includegraphics[width=\linewidth]{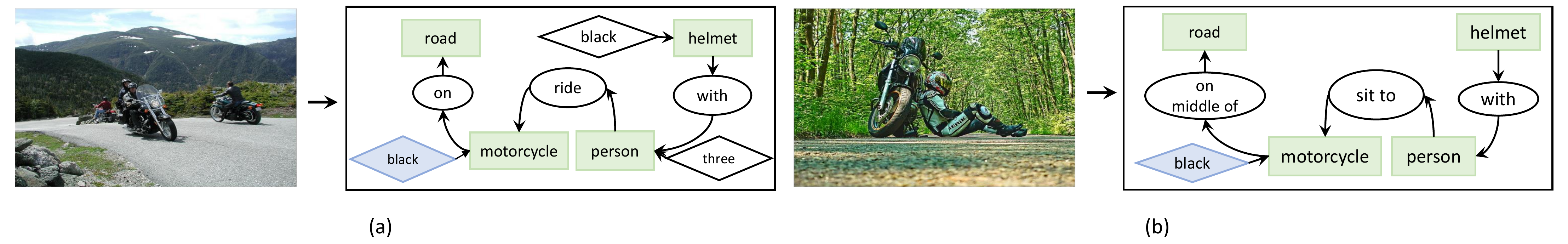}
\caption{An example of parsed scene graphs for the ground-truth captions of two images. Two graphs have four object overlaps: ``\texttt{helmet}'', ``\texttt{people}'', ``\texttt{motorcycle}'', and ``\texttt{road}'' (\textcolor{mygreen}{green}), and one attribute overlap ``\texttt{black}'' (\textcolor{myblue}{blue}).}
\label{fig_3_fine_grained_matching}
\end{figure}

\addtolength{\tabcolsep}{-1pt}
\begin{table}[t]
  \centering

    \caption{Statistical summary of the COCO-DIC, Flickr30K-DIC, and existing Ref-DIC benchmark~\cite{wang2021group}. ``\#overlaps'' denotes the number of object/attribute overlap in each dataset.}
    \begin{tabular}{l|cccc|ccc}
    \hline
    \multirow{2}[2]{*}{Datasets} & \multicolumn{4}{c|}{images} & \multicolumn{3}{c}{\#overlaps in a group}\\
    \multicolumn{1}{c|}{} & 
    \multicolumn{1}{c}{Train} & \multicolumn{1}{c}{Val} & \multicolumn{2}{c|}{Test} & \multicolumn{1}{c}{Train} & \multicolumn{1}{c}{Val} & \multicolumn{1}{c}{Test} \\
    \hline
    Wang~\etal~\cite{wang2021group} & 133,980 & 5,562  & \multicolumn{2}{c|}{5,538} & 3.8   & 3.7   & 3.7 \\
    COCO-DIC & 123,287 & 5,000  & \multicolumn{2}{c|}{5,000} & 5.0     & 4.9   & 4.9 \\
    Flickr30K-DIC & 29,000 & 1,014  & \multicolumn{2}{c|}{1,000} & 5.3   & 5.3   & 5.3 \\
    \hline
    \end{tabular}%
 
   \label{table:dataset_statics}%
\end{table}%
\addtolength{\tabcolsep}{1pt}

\subsubsection{Fine-grained Group Construction}
To overcome the shortcoming of the coarse-grained group, we propose a fine-grained matching mechanism that directly uses object and attribute overlaps between two images as the similarity measurement. Firstly, for $I_t$ or any image in $\mathcal{I}_c$, we parse all its ground-truth captions into one scene graph. Then, we extract objects and attributes from scene graphs~\cite{chen2019counterfactual,xu2020scene,li2022devil,liu2021toward} of two images to calculate overlaps. Specifically, objects from the two graphs will be compared according to their categories. However, two attributes should first correspond to the same objects and then compare to each other. For example in Fig.~\ref{fig_3_fine_grained_matching}, when calculating object overlaps, ``\texttt{black helmet}'' (top) and ``\texttt{helmet}'' (bottom) from two graphs denote one-time object overlap (\eg, common object ``\texttt{helmet}''). As for attribute overlaps, two graphs have  both ``\texttt{black motorcycle}'' in common and denote one-time attribute overlap\footnote{Note that the ``\texttt{black helmet}'' and  ``\texttt{black motorcycle}'' contain no attribute overlap because the attribute ``\texttt{black}'' belongs to different objects.}. In this paper, we take the sum over object and attribute overlaps as the final similarity score for two images. And we sort all images in $\mathcal{I}_c$ according to their similarity scores to $I_t$ to construct the fine-grained group $\mathcal{I}_f$. It is worth noting that we select $K$ images but not the top-K from $\mathcal{I}_f$ to construct $I_r$. The reason for this choice is that we believe the most similar images from $\mathcal{I}_f$ may contain the identical objects and attributes as $I_t$, thus they won't help to emphasize any unique details in $I_t$. More detailed discussions about the top-K selection are left in Table~\ref{table:abal_group}.

\subsection{Benchmarks: COCO-DIC \& Flickr30K-DIC}
We apply our matching mechanism to widely-used captioning benchmarks MS-COCO~\cite{2015Microsoft} and Flickr30K~\cite{2016Flickr30k} to construct \textbf{COCO-DIC} and \textbf{Flickr30K-DIC}, respectively. Some basic statistics about our proposed benchmarks are reported in Table~\ref{table:dataset_statics}. Different from the construction procedure proposed in~\cite{wang2021group}, they avoid image reuse (or overlap) among different constructed groups. Thus, some images which are not similar enough may be forced to construct a group. In contrast, we find $K$ reference images independently to ensure similarity within a group.

\section{Proposed Approach}

\subsection{Preliminaries}
\subsubsection{Transformer-based Image Captioning}
\label{sec:transformer IC}
Transformer~\cite{vaswani2017attention} follows the standard \textbf{Encoder-Decoder} architecture. It employs the self-attention mechanism to explore the internal correlation within the sequential data, which has been widely adopted by numerous image captioning models~\cite{li2019entangled,2020Meshed}. 
For a given image $I$, they use proposal features~\cite{anderson2018bottom} extracted by an object detector as input: $X=\{x_i\}_{i=1}^N$ where $x_i \in \mathbb{R}^d$ is the feature vector for $i$-th proposal in $I$ and $N$ is the number of proposals. They employ multiple self-attention layers as \textbf{Encoder}, and the outputs of the $l$-th layer are calculated as follows:
\begin{align} \label{self-attn layers}
		H_{l-1} & =\textbf{LN}\left( O_{l-1}+\textbf{MH}( O_{l-1}, O_{l-1}, O_{l-1}) \right), \\
		O_{l}  & = \textbf{LN}\left( H_{l-1} + \textbf{FFN}(H_{l-1}) \right),
\end{align}
\noindent where $O_0$ refers to input proposal features $X$, and $O_{l-1}$ is outputs of the $(l-1)$-th layer. $\textbf{LN}(\cdot)$ denotes the layer normalization~\cite{2016Layer}, $\textbf{FFN}(\cdot)$ denotes the feed forward network, and $\textbf{MH}(\cdot)$ denotes the multi-head attention~\cite{vaswani2017attention}. Encoded visual features are fed into the \textbf{Decoder} for caption generation. It generates a word probability distribution $P_t = P(w_t|w_{1:t-1},I)$ at each time step t conditioning on the previously generated words $\{w_1,\ldots,w_{t-1}\}$ and image $I$.

\subsubsection{Model Optimization}
Mainstream captioning works typically resort to a two-stage procedure for model optimization~\cite{rennie2017self,2020Meshed}. Given an image $I$ and its ground-truth captions $C = \{c_i\}_{i=1}^M$. They first apply a cross-entropy loss (XE) to pre-train the model and then employ reinforcement learning (RL) to finetune sequence generation. 

When training with XE, for a ground-truth caption $c_i = \{w_t\}_{t=1}^T$, they ask the model to minimize the following cross-entropy loss:
\begin{equation}
    L_{xe} = - \sum_{t=1}^T\log P(w_t|w_{1:t-1},I),
\end{equation}
where $P(w_t|w_{1:t-1},I)$ denotes the predicted probability of word $w_t$.

When training with reinforcement learning, they firstly generate top-n captions $\hat{C} = \{\hat{c}_i\}_{i=1}^n$ through beam search, and then optimize the following RL loss~\cite{2020Meshed}\footnote{Note that there are several RL variants for caption generation, we only demonstrate one typical policy gradient solution here.}:
\begin{equation}
    \label{eq:4}
    L_{rl} = - \frac{1}{n} \sum_{i=1}^n ((r(\hat{c}_i, C) - b) \log p(\hat{c}_i)),
\end{equation}
where r($\cdot,\cdot$) is the reward function computed between $\hat{c}_i$ and $C$, and $b = (\sum_{i=1}^n r(\hat{c}_i,C)) / n$ is the baseline, calculated as the mean of the rewards obtained by the generated captions.

\begin{figure*}[t]
\centering
\includegraphics[width=\linewidth]{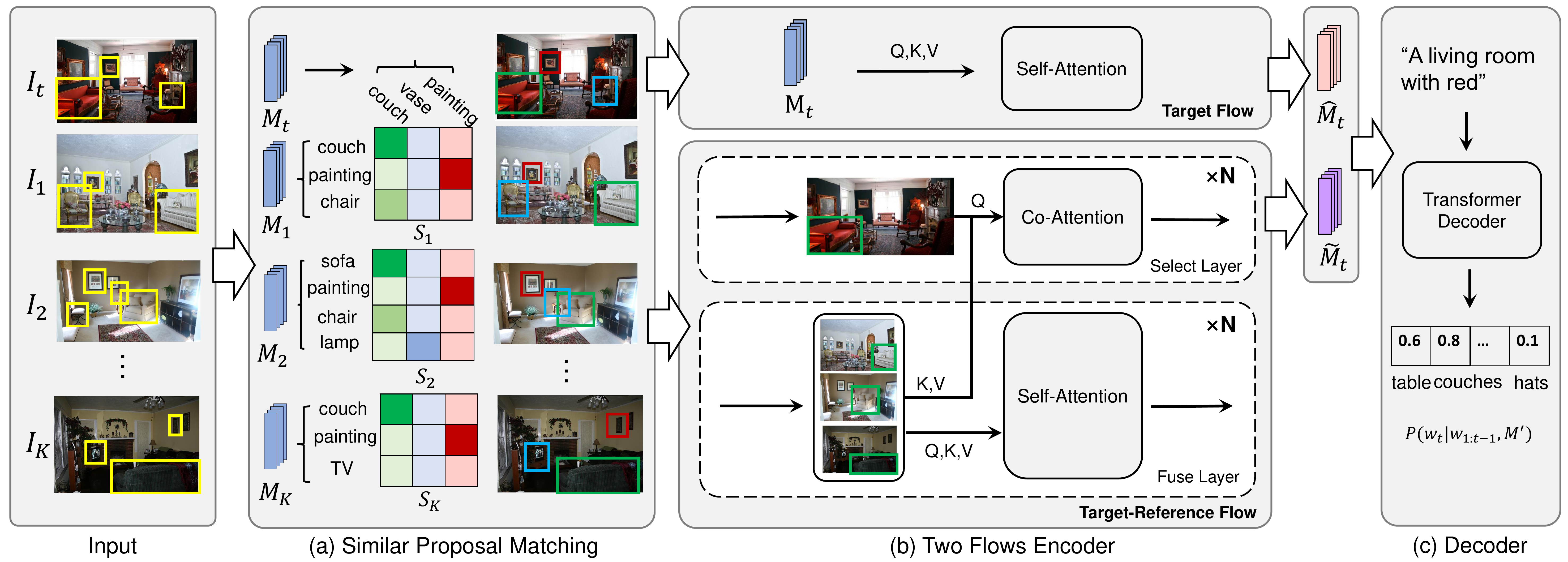}
\caption{Overview of our proposed TransDIC model. It consists of three parts: (a) A Similar Proposal Matching module that couples each target proposal with reference proposals. (b) A Two-Flow Encoder module that encodes both target and reference images. (c) A plain captioning decoder. In module (a), we use the same colors to denote the same objects. We first send target and reference images into the similar proposal matching module to construct Target-Reference tuples. Then we send constructed tuples into the Two-Flow Encoder for target image and cross-image features extraction. Finally, both kinds of features are sent into the decoder for caption generation.}
\label{fig_4_model_overview}
\end{figure*}

\subsection{TransDIC: Transformer-based Ref-DIC}
Given a target image $I_t$, we term all regions (or proposals) in it as \textbf{target proposals} $R^t=\{r_n^t\}_{n=1}^N$. Our proposed model tends to give each target proposal $r_n^t$ some proposal references when generating captions. To this end, we first couple each $r_n^t$ with semantic-similar proposals from $\mathcal{I}_r$ as the reference, \ie, \textbf{reference proposals}. Then we send each target proposal and its reference proposals into the model for distinctive caption generation.

Specifically, our TransDIC consists of three components: 
1) Similar Proposal Matching module in Section~\ref{sec: sim_box_matching}. 2) Two-Flow Encoder module. The module contains two parallel data flows to extract both target image and cross-image features in Section~\ref{sec:select-fuse}. 
3) A plain Transformer-based Captioning Decoder for caption generation. An overview of our model is shown in Fig.~\ref{fig_4_model_overview}.

\subsubsection{Similar Proposal Matching} \label{sec: sim_box_matching}
For each target proposal, we retrieve proposals from reference images with the highest similarity scores to it as its reference proposals. Given a target image $I_t$ and its corresponding $\mathcal{I}_r$, their proposal features are firstly projected into the memory space through an MLP layer. We
denote memory features for $I_t$ and the K images in $\mathcal{I}_r$ as $M_t =\{m^t_j\}_{j=1}^N$ and $M_{k}=\{m^k_i\}_{i=1}^N,k=\{1\dots K\}$ , respectively.

Then, we calculate the cosine similarity scores between features in $M_t$ and $M_k$, \ie, 
\begin{equation}
    S^k_{ij} = cos(m^k_{i},m^t_j),
\end{equation}
where $m^t_j$ represents the $j$-th proposal in $I_t$, and $m^k_{i}$ represents the $i$-th proposal in reference image $I_k$. We apply max operation to get the most similar proposal for $r_j^t$ according to the calculated $S^k$:
\begin{equation}
    \hat{r}^k_j = \mathop{\arg\max}_i(\{S^k_{ij}\}_{i=1}^N),
\end{equation} 
where $\hat{r}^k_j$ denotes the most similar proposal from image $I_k$ for $r_j^t$, \ie, reference proposal.
As an example in Fig.~\ref{fig_4_model_overview}(a), reference image $I_K$ contains proposals ``\texttt{couch}'', ``\texttt{painting}'' and ``\texttt{TV}''. For the proposal ``\texttt{couch}'' in $I_t$, we can learn from the similarity matrix $S^K$: the proposal is similar to the ``\texttt{couch}'' in $I_K$ (deep green) while is different from the ``\texttt{TV}'' or ``\texttt{painting}'' (light green) in $I_K$. Max operation is then token along each column of $S^K$, and ``\texttt{couch}'' in $I_k$ is selected as the reference proposal for ``\texttt{couch}'' in $I_t$.

Finally, for each target proposal $r_n^t$, we gather $K$ reference proposals, one for each, from $K$ reference images. We put these $K+1$ proposals together and term them as a \textbf{Target-Reference proposal tuple}:
\begin{equation}
    T_n = \{r_n^t,\hat{r}_n^1,\hat{r}_n^2,\dots,\hat{r}_n^K \}. \quad n= \{1, \dots ,N \}
\end{equation}
For example in Fig.~\ref{fig_4_model_overview}(a), all proposals marked with green boxes form a Target-Reference tuple (for ``\texttt{couch}'').

\subsubsection{Two-Flow Encoder} \label{sec:select-fuse}
Our proposed  module takes $M_t$ and $M_k$ as input, and extracts target image features and cross-image features through the \textbf{Target flow} and the \textbf{Target-Reference flow},  respectively. An overview of the module is shown in Fig.~\ref{fig_4_model_overview}(b).

\textbf{Target flow.} The flow enables the proposal feature interactions within the target image $I_t$. Same as the standard transformer-based captioning model, it sends memory features $M_t$ into multiple self-attention layers, and finally outputs encoded features $\hat{M}_t=\{\hat{m}^t_i\}_{i=1}^N$ for $I_t$.

\textbf{Target-Reference flow.} This data flow consists of \textbf{select layers} and \textbf{fuse layers}. Given a Target-Reference proposal tuple $T_n$, we denote the memory features for target proposal and reference proposals in $T_n$ as $\bar{m}_n^t$ and $\bar{M}_n = \{\bar{m}_n^i\}_{i=1}^K$, respectively. The flow takes in those two kinds of features and generates cross-image features through \textbf{select} and \textbf{fuse} layers:

\textbf{\emph{Fuse layer}}. The goal of the fuse layer is to enable the interactions among memory features within $\bar{M}_n$. We stack multiple fuse layers, and the $l$-th fuse layer is calculated as follows:
\begin{equation}
    U_{l} = \textbf{MH}(U_{l-1},U_{l-1} ,U_{l-1}),
\end{equation}
where $U_{0}$ refers to $\bar{M}_n$ and $U_{l-1}$ is the outputs of the $(l-1)$-th fuse layer. 
Because all the features in $\bar{M}_n$ are semantic-similar, the model can learn to capture the primary concepts they are describing. 

\textbf{\emph{Select layer}}. The select layer builds on the co-attention mechanism. We set the features of the target proposal as the Query, the features of reference proposals as the Key and Value in multi-head attention. Multiple co-attention layers are stacked, and the $l$-th select layer is computed as:
\begin{equation}
    	V_{l} = \textbf{MH}(V_{l-1},U_{l-1} ,U_{l-1}),
\end{equation}
where $V_{0}$ refers to $\bar{m}_n^t$, $V_{l-1}$ and $U_{l-1}$ are the outputs of the $(l-1)$-th select and fuse layer, respectively. By the residual connection in self-attention blocks, feature $\bar{m}_n^t$ will gradually select useful information from reference images while preserving the original information from $I_t$.

As an example shown in Fig.~\ref{fig_4_model_overview}(b), our model can learn to focus on the unique attributes and objects in $I_t$. For unique attributes, we send all reference proposals of ``\texttt{couch}'' (green boxes) into fuse layers, the model will be informed they are describing the concept ``\texttt{couch}''. Since the target proposal also describes ``\texttt{couch}'', the select layer learns to focus on the unique color ``\texttt{red}'' of the ``\texttt{couch}'' in $I_t$. When predicting unique objects, for ``\texttt{vase}'' proposal in $I_t$, because all selected reference proposals for it (blue boxes) do not contain the same concept, the select layer learns that ``\texttt{vase}'' is a unique object in $I_t$.

We use the outputs of the last select layer as the final refined target feature $\tilde{m}_n^t$ for proposal $r_n^t$ in $I_t$. For N Target-Reference tuples in $I_t$, we can get $\tilde{M}_t =  \{\tilde{m}_i^t\}_{i=1}^N$ as the outputs of Target-Reference flow. Finally, we concatenate the outputs of the Target flow and Target-Reference flow as the final outputs of the Two-Flow Encoder:
\begin{equation}
    M_t' = [\hat{M}_t;\tilde{M}_t],
\end{equation}
where $[\cdot;\cdot]$ denotes concatenation operation, and $M_t'$ will be sent into decoder for caption generation. 

\subsection{\textbf{TransDIC++}: TransDIC with DisReward}
\label{sec:contrastive}
Given a target image $I_t$ and its corresponding reference images $\mathcal{I}_r$, using the  notation mentioned in    sec.\ref{sec:transformer IC}, we construct the positive sample with proposal features of target and reference images $X_{pos}=\{X_t,\mathcal{X}_r\}$, while the negative sample is constructed by masking some proposals in the reference images to zeros $X_{neg}=\{X_t, \mathcal{X}_r^{mask}\}$. By removing some information in the reference images, the generated caption for the target image should be worse compared to the caption generated from the positive sample.

Specifically, to enforce the comparison between positive and negative samples, we design a new reward named \textbf{DisReward} at the reinforcement learning stage. Meanwhile, two kinds of masking strategies: 1) instance-level masking; 2) image-level masking are designed to study the influence of masking strategy.  We will first introduce the overall architecture of the module in \ref{sec:disreward} and then explain the details of the masking strategy in Sec.~\ref{sec:instance} and Sec.~\ref{sec:image}.

\subsubsection{Contrastive Learning Architecture: Overall pipeline}
\label{sec:disreward}
Our designed architecture can be easily adopted by any kinds of Ref-DIC models. Specifically, both positive sample and negative sample are sent into the captioning model, which we denote as $f(X)$, for caption generation:
\begin{align}
    c_{pos} = f(X_{pos}), \\
    c_{neg} = f(X_{neg}).
\end{align}

\begin{figure*}[t]
\centering
\includegraphics[width=0.9\linewidth]{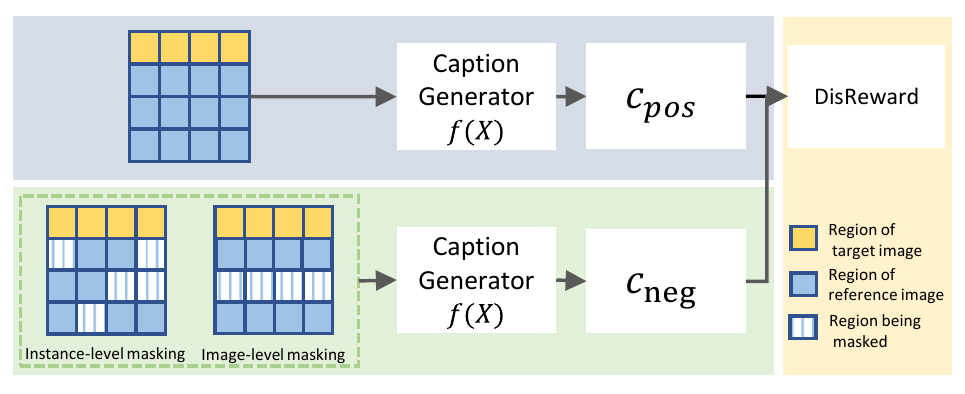}
\caption{Overview of our proposed CL module. It consists of two kinds of flows: positive flow and negative flow. Positive and negative samples are sent into the flow and generate $C_{pos}$ and $C_{neg}$, respectively. Finally, these two generated captions are used to calculate DisReward.}
\label{fig_contrastive}
\end{figure*}

\noindent\textbf{DisReward}. We design a new reward named DisReward to encourage the model to generate more distinctive captions, it enforces the caption generated by positive samples has a higher evaluation score than the ones generated by negative samples. Specifically, we firstly use BLEU~\cite{papineni2002bleu} and CIDEr~\cite{vedantam2015cider} to construct a mixed metric \textbf{bleuder} as the final evaluation metric. For any generated caption $c$, bleuder is calculated as:
\begin{equation}
    bleuder(c) = \alpha_b \times BLEU_{1}(c) + \alpha_b \times BLEU_{4}(c) + \alpha_c \times CIDEr(c).
\end{equation}
$\alpha_b,\alpha_c$ is the hyperparameter control the coefficient of the metrics.

Then, DisReward is calculated as:
\begin{equation}
    DisReward(c_{pos}) = -\mathop{max}(0,bleuder(c_{neg}) - bleuder(c_{pos}) + \beta),
\end{equation}
\noindent where $\beta$ is the hyperparameter margin. Intuitively, $DisReward$ ``wants'' the model to assign a higher reward to positive samples (by at least $\beta$) than negative samples because some proposals are masked to zeros and the target image ``gets less information'' from reference images.

Finally, following~\cite{luo2018discriminability}, the $DisReward$ is used as an additional reward and the final reward for the reinforcement learning is:
\begin{equation}
    r(c_{pos},C) = CIDEr(c_{pos}) + \lambda \times DisReward(c_{pos}).
\end{equation}
\noindent Where $\lambda$ is a hyperparameter. Reward function in Eq.~\eqref{eq:4} is replaced with this new reward for loss backpropagation.

\subsubsection{Instance-level Masking}
\label{sec:instance}

Instance-level masking masks out some proposals in the reference images. For each proposal in the target image, we aim to mask those strongly related proposals in the reference images because they may contribute more to the caption generation. We believe that these samples contain ``less information'' can be served as negative samples.

\noindent \textbf{Similarity Matrix.} In sec.\ref{sec: sim_box_matching}, we have already calculated the similarity between proposals in target and reference images, we directly reuse the calculated similarity matrix. Specifically, $X_{pos}$ is firstly sent to the caption generator, through which, we can get the similarity matrix $S_{ij}^k$. As we want to remove useful information in the reference images, we mask out proposals that have higher scores than proposals in the target image. Specifically, if the similarity between a target proposal and a reference proposal is larger than the threshold, this reference proposal will be masked. The generated mask will be applied to $\mathcal{X}_r$ to construct $\mathcal{X}_r^{\text{mask}}$.

\noindent \textbf{Grad-CAM.} Following~\cite{gao2022open}, we use grad-cam~\cite{selvaraju2017grad} to investigate the importance of proposals in reference images. Intuitively, if the proposal contributes more to the caption generation, it will receive more gradients. Specifically, we firstly use a pretrained multi-modality model UNITER\cite{chen2020uniter} to calculate the similarity $\textbf{s}$ between generated $C_{pos}$ and $\mathcal{X}_r$. Then we calculate the gradient of $\textbf{s}$ with respect to $\mathcal{X}_r$:
\begin{equation}
    \Phi = \frac{\partial \textbf{s}}{\partial \mathcal{X}_r}.
\end{equation}
We sort the calculated gradient to get those more important proposals. Obviously, if the gradients are large, then these proposals are more important to caption generation and will be masked. We set a pre-defined threshold and $cumsum$ all gradient scores in reverse manner, proposals within the threshold are masked. Finally, the generated mask is applied to $\mathcal{X}_r$ to construct $\mathcal{X}_r^{mask}$.

\subsubsection{Image-level Masking}
\label{sec:image}
Image-level masking masks an entire image or substitute some reference images with other unrelated images. This coarse-grained strategy  masks out ``more negative'' samples, we still aim to remove those strongly related images.

\noindent \textbf{Image Pool.}
To remove strongly related images in reference images, we build up an image pool. As mentioned in sec.\ref{sec:sim_group}, we select top-N (N > K) images from the fine-grained group to construct the pool. For each forward pass, we randomly sample K images from the pool as final reference images. In this manner, we construct $\mathcal{X}_r^{mask}$ in a more ``soft'' way, and more information is left over.

\noindent \textbf{Grad-CAM.}
Similar to the method in instance-level masking, we use UNITER to calculate the similarity between $C_{pos}$ and all reference images one by one. We then task sum over all proposals within each individual image as its final score. Images that get higher scores will be masked entirely.
\begin{equation}
    \Phi^i = \frac{\partial \textbf{s}}{\partial \mathcal{X}_r^i} \quad i \in \{1,\dots,K\}.
\end{equation}

\section{Experiments}
In this section, we describe the datasets used for experiments and introduce a new distinctiveness-based evaluation metric \textbf{DisCIDEr}. We conduct extensive experiments and ablation studies to reveal the superiority of our proposed model, and our proposed benchmarks for Ref-DIC.

\subsection{Datasets}
We developed the \textbf{COCO-DIC} and \textbf{Flickr30K-DIC} based on the MS-COCO~\cite{2015Microsoft} and Flickr30K~\cite{2016Flickr30k}. They contain 123287 and 31014 images,  respectively. Each image is annotated with 5 ground-truth captions. For both datasets, we followed the splits provided by~\cite{karpathy2015deep}, and constructed reference image groups within the training, validation, and test splits. For completeness, we also reported results on Wang~\etal~\cite{wang2021group}'s dataset for Ref-DIC.

\subsection{Evaluation Metrics}
We applied two kinds of metrics to evaluate the accuracy and distinctiveness of generated captions. For accuracy evaluation, we calculated four commonly used evaluation metrics: BLEU-N (B-N) (1- to 4-grams)~\cite{papineni2002bleu}, ROUGE-L (R)~\cite{lin2004rouge}, METEOR (M)~\cite{banerjee2005meteor}, and CIDEr (C)~\cite{vedantam2015cider}. For distinctiveness evaluation, we developed a new metric named \textbf{DisCIDEr} (DisC). We introduce the DisCIDEr below.

\noindent\textbf{DisCIDEr.} All existing metrics designed for the Ref-DIC task fail to fully explore the distinctiveness of each individual n-gram in GT captions of the target image. To solve this, we assign these n-grams with different weights according to group-level distinctiveness: if an n-gram occurs frequently in ground-truth captions of reference images, it is less distinctive. As an example in Fig.~\ref{fig_5_discider_example}, both target and reference images are describing ``\texttt{red sofa}'', so we should assign lower weights to the word ``\texttt{red}'' in ground-truth captions of $I_t$ at evaluation time. Instead, since only image $I_t$ contains the object ``\texttt{fireplace}'', we should put more weight on it.

To realize this intuition, we modify the n-gram weighting procedure of CIDEr by adding a re-weight term. For a target image $I_t$, we denote its generated and ground-truth captions as $c$ and  $S_t=\{s_{t}^i\}_{i=1}^M$, respectively. Similarly, we denote ground-truth captions for reference images $\mathcal{I}_r$ as $S_r=\{s_{r}^j\}_{j=1}^M,r=\{1\ldots K\}$. The number of times an n-gram $\omega_d$ occurs in $s_{t}^j$ is denoted as $h_d(s_{t}^j)$. We modify CIDEr by adding an \textbf{Inverse reference frequency} term after it when calculating $g_d(s_{t}^j)$ for the n-grams in ground-truth captions of $I_t$:

\begin{figure}[t]
\centering
\includegraphics[width=\linewidth]{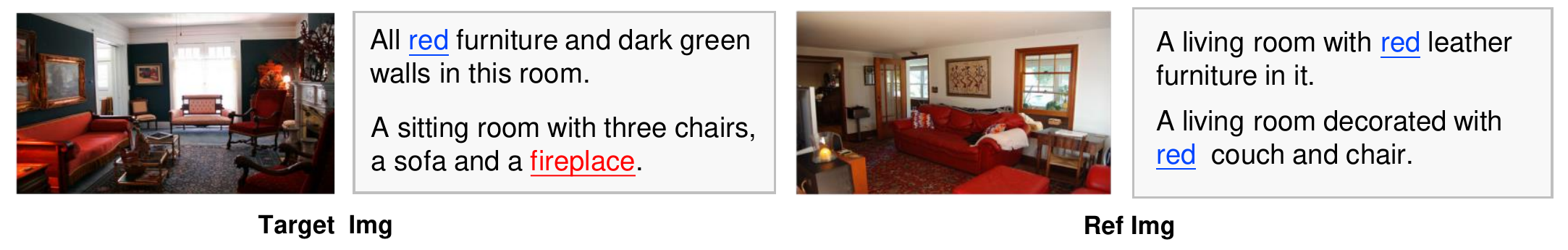}
\caption{An example of the intuition behind DisCIDEr. N-gram ``\texttt{red}'' appears in both target and reference images, thus should be given less attention (\textcolor{myblue}{blue}). In contrast, we should pay more attention to ``\texttt{fireplace}'', because it  appears only in the target image (\textcolor{red}{red}).}
\label{fig_5_discider_example}
\end{figure}

\begin{small}
\begin{align}
    g_d(s_{t}^i) =  
    \frac{h_d(s_{t}^i)}{\sum_{w_l\in\Omega} h_l(s_{t}^i)}   
    \underbrace{
    \log (\frac{|I|}{\max (1, \sum_{I_p \in I} \min(1,\sum_{s_p^q  \in I_p} h_d(s_{p}^q)))}) }_{\text{Inverse document frequency}}
    {\underbrace{
    \log (\frac{m + K}{n+\sum_{S_u \in {S_{1:K}}} \min (1,\sum_{s_{u}^v\in S_u} h_d(s_{u}^v))})   
    }_{\textcolor{mygreen}{\text{Inverse reference frequency}}}},
\end{align}
\end{small}    
where $\Omega$ is the vocabulary of all n-grams and $I$ is the set of all images. $m$ and $n$ are two parameters. In this way, DisCIDEr can evaluate the group-level distinctiveness while preserving the advantages of n-gram-based metrics. We refer the readers to~\cite{vedantam2015cider} for more details.

\subsection{Implementation Details}
Following~\cite{anderson2018bottom}, we used the region proposal features extracted by the Faster R-CNN~\cite{ren2016faster} with dimension 2048, and the memory space was of dimension 512. The number of self-attention blocks in Target flow, select and fuse layers in Target-Reference flow were set to 3. In the two-stage matching procedure, the size of $\mathcal{I}_c$ was 500. For $\mathcal{I}_r$, we used images with top-p to top-(p+K-1) highest similarity scores, where p is an adjustable parameter for group similarity and both benchmarks set p to 3, K to 5. Parameters $m$ and $n$ in DisCIDEr were set to 0.8 and 5.0.


\subsection{Comparison with State-of-the-Art Methods}
We reported our results on two kinds of datasets: 1) Our constructed COCO-DIC and Flickr30K-DIC datasets. 2) The dataset proposed in~\cite{wang2021group}. We compared our TransDIC model with three kinds of state-of-the-art models: 1) \textbf{NIC}~\cite{vinyals2015show}, \textbf{Xu~\etal}~\cite{xu2015show}, \textbf{UpDown}~\cite{anderson2018bottom}, 
\textbf{AoANet}~\cite{huang2019attention}, 
\textbf{Transformer}~\cite{vaswani2017attention}, \textbf{$M^2$ Transformer}~\cite{2020Meshed}. They only aim to generate captions with high accuracy. 2) \textbf{DiscCap}~\cite{luo2018discriminability}, \textbf{CIDErBtwCap}~\cite{wang2020compare}. They are designed for the Single-DIC. 3) \textbf{GdisCap}~\cite{wang2021group} that is designed for the Ref-DIC. We also compared \textbf{CAGC}~\cite{li2020context} which use multiple images as input.


\noindent\textbf{Results on COCO-family Benchmarks.} From Table~\ref{table:result_on_coco}, we can observe: 1) For accuracy evaluation, our proposed \textbf{TransDIC++} achieves the best performance on all metrics at both COCO-DIC and Wang~\etal~\cite{wang2021group} (\eg, 132.4 v.s. 125.8 in GdisCap on CIDEr). Meanwhile, our model outperforms some strong state-of-the-art models (\eg, 132.4 v.s. 131.2 in $M^2$ Transformer on CIDEr) in terms of accuracy-based metrics. 2) For distinctiveness evaluation, our model gets the highest scores on DisCIDEr in the two datasets. 3) We also adapted our proposed \textbf{DisReward} to existing Ref-DIC model GdisCap, as shown in the Table~\ref{table:result_on_coco}~(row 10), the \textbf{DisReward} can steadily improve the performance of the existing model, demonstrating the effectiveness of the reward.

\noindent\textbf{Results on Flickr30K-family Benchmarks.} From Table~\ref{table:result_on_flickr}, we can observe: 1) For accuracy evaluation, our \textbf{TransDIC++} achieves the largest performance gains on all the metrics especially on CIDEr (69.1 v.s. 65.6). 2) For distinctiveness-based metrics, our model outperforms GdisCap and TransDIC by a large gap on DisCIDEr. 
\begin{table}[t]

    \caption{ Comparison of captions accuracy on COCO family with state-of-the-art image captioning models. }
    \centering
  
    \begin{tabular}{lcccccc}
    \hline
    Model & B-1 & B-4 & M & R & C & DisC \\
    \hline    
        \multicolumn{7}{l}{\textbf{Dataset: MS-COCO}} \\
        UpDown~\cite{anderson2018bottom}  & 79.8  & 36.3  & 27.7  & 56.9  & 120.1 & \multicolumn{1}{c}{---}  \\
        AoANet~\cite{huang2019attention}  & 80.2  & 38.9  & 29.2  & 58.8  & 129.8 & \multicolumn{1}{c}{---}  \\
        Transformer~\cite{vaswani2017attention} & 80.0 & 38.2  & 28.9 & 58.2 & 127.3 & 98.7 \\
        $M^2$ Transformer~\cite{2020Meshed}   & 80.8  & 39.1  & 29.2  & 58.6  & 131.2 & \multicolumn{1}{c}{---}  \\
    \cdashline{1-7}[1pt/1pt]
        DiscCap~\cite{luo2018discriminability} & \multicolumn{1}{c}{---}   & 36.1  & 27.4  & 57.3  & 114.3 & \multicolumn{1}{c}{---} \\
        CIDErBtwCap~\cite{wang2020compare} &  \multicolumn{1}{c}{---}   & 38.5  & 29.1  & 58.8  & 127.8 & \multicolumn{1}{c}{---} \\
    \hline
        \multicolumn{7}{l}{\textbf{Dataset: COCO-DIC}} \\
        GdisCap~\cite{wang2021group} & 80.0 & 37.3 & 28.4 & 57.5 & 125.8 & 96.6  \\
        GdisCap+DisReward~ & 80.8 & 38.2 & 28.7 & 58.1 & 127.0 & 98.1  \\
        \cdashline{1-7}[1pt/1pt]
        CAGC~\cite{li2020context}  & 80.7 & 38.1 & 28.7 & 57.9 & 127.9 & 98.0 \\
        \cellcolor{mygray-bg}{\textbf{TransDIC (Ours)}} & \cellcolor{mygray-bg}{81.6} & \cellcolor{mygray-bg}{39.3} & \cellcolor{mygray-bg}{29.2} & \cellcolor{mygray-bg}{58.5} & \cellcolor{mygray-bg}{132.0} & \cellcolor{mygray-bg}{102.2} \\
        \cellcolor{mygray-bg}{\textbf{TransDIC++ (Ours)}} & \cellcolor{mygray-bg}{\textbf{82.7}} & \cellcolor{mygray-bg}{\textbf{40.3}} & \cellcolor{mygray-bg}{\textbf{29.1}} & \cellcolor{mygray-bg}{\textbf{58.9}} & \cellcolor{mygray-bg}{\textbf{132.4}} & \cellcolor{mygray-bg}{\textbf{102.4}} \\
    \hline
        \multicolumn{7}{l}{\textbf{Dataset: Wang~\etal~\cite{wang2021group}}} \\
        GdisCap~\cite{wang2021group}  &  80.2 & 37.7  &  28.3 &   57.3 & 126.6 & 97.7 \\
        CAGC~\cite{li2020context}  &  80.4 & 37.7  & 28.7  & 57.6   &  127.2 & 98.0 \\
        \cellcolor{mygray-bg}{\textbf{TransDIC (Ours)}} & \cellcolor{mygray-bg}{81.0} & \cellcolor{mygray-bg}{38.8} & \cellcolor{mygray-bg}{\textbf{29.1}} & \cellcolor{mygray-bg}{58.2} & \cellcolor{mygray-bg}{130.8} & \cellcolor{mygray-bg}{101.9} \\
        \cellcolor{mygray-bg}{\textbf{TransDIC++ (Ours)}} & \cellcolor{mygray-bg}{\textbf{82.0}} & \cellcolor{mygray-bg}{\textbf{39.7}} & \cellcolor{mygray-bg}{29.0} & \cellcolor{mygray-bg}{\textbf{58.7}} & \cellcolor{mygray-bg}{\textbf{131.3}} & \cellcolor{mygray-bg}{\textbf{102.0}} \\
    \hline
    
    \end{tabular}%
    
     \label{table:result_on_coco}
\end{table}%

\begin{table}[t]
    \caption{ Comparison of captions accuracy on Flickr30K family with state-of-the-art image captioning models. }
    \centering

    \begin{tabular}{{lcccccc}}
    \hline
    Model & B-1 & B-4 & M & R & C & DisC\\
    \hline    
        \multicolumn{7}{l}{\textbf{Dataset: Flickr30K}} \\
        NIC~\cite{vinyals2015show} & 66.3  & 18.3 & --- & --- & --- & ---  \\
        Xu~\etal~\cite{xu2015show} & 66.9 & 19.9 & 18.5 & --- & --- & ---  \\
        Transformer~\cite{vaswani2017attention} & 70.7 & 27.7  & 21.4 & 49.0 & 61.2 &39.1  \\
    \hline
        \multicolumn{7}{l}{\textbf{Dataset: Flickr30K-DIC}} \\
        GdisCap~\cite{wang2021group} & 71.7 & 29.0 & 22.1 & 49.6 & 65.6 & 41.2 \\
        CAGC~\cite{li2020context}  & 72.9 & 29.1 & 21.9 & 50.1 & 62.2 &39.0  \\
        \cellcolor{mygray-bg}{\textbf{TransDIC}}  & \cellcolor{mygray-bg}{73.2} & \cellcolor{mygray-bg}{30.1} & \cellcolor{mygray-bg}{22.5} & \cellcolor{mygray-bg}{50.3} & \cellcolor{mygray-bg}{65.1} & \cellcolor{mygray-bg}{41.4} \\
        
        \cellcolor{mygray-bg}{\textbf{TransDIC++ (Ours)}} & \cellcolor{mygray-bg}{\textbf{74.4}} & \cellcolor{mygray-bg}{\textbf{31.1}} & \cellcolor{mygray-bg}{\textbf{23.5}} & \cellcolor{mygray-bg}{\textbf{51.4}} & \cellcolor{mygray-bg}{\textbf{69.1}} & \cellcolor{mygray-bg}{\textbf{44.4}} \\
    \hline
    \end{tabular}%

    \label{table:result_on_flickr}
\end{table}%

\addtolength{\tabcolsep}{1pt}
\begin{table}[t]
  \centering
    \caption{Ablation study of Two-Flow Encoder on COCO-DIC. ``Fuse'' and ``Select'' denote the fuse layer and select layer in the Target-Reference flow, respectively.}
    \large
    \begin{tabular}{cc | cccccc}
    \hline
    Fuse & Select & \multicolumn{1}{c}{B-1} & \multicolumn{1}{c}{B-4} & \multicolumn{1}{c}{M} & \multicolumn{1}{c}{R} & \multicolumn{1}{c}{C} & \multicolumn{1}{c}{DisC} \\
    \hline
    \ding{55} & \ding{55} & 80.0    & 38.2  & 28.9 & 58.2 & 127.3 & 98.7 \\
    \ding{55} & \ding{51} & 81.3  & 38.4  & 29.0    & 58.1  & 130.0   & 100.2 \\
    \ding{51} & \ding{55} &  \textbf{81.6}   & 39.1  & \textbf{29.1}  & 58.4  & 131.3 & 101.6  \\
    \ding{51} & \ding{51}  & \textbf{81.6}  & \textbf{39.3}  & \textbf{29.1}  & \textbf{58.5}  & \textbf{132.0}   & \textbf{102.2} \\
    \hline
    \end{tabular}%

    \label{table:abal_module}
\end{table}%
\addtolength{\tabcolsep}{-1pt}

\subsection{Ablation Studies}
We conducted extensive experiments to verify the influences of the proposed Two-Flow Encoder module and group similarity.

\subsubsection{Influence of Two-Flow Encoder}
To measure the influence of each component in our proposed Target-Reference flow, we trained an ordinary transformer as the baseline and three variants of our model. 1) Target-Reference flow only contains the select layer: stacked select layers always take the original reference features as input. 2) Target-Reference flow only contains the fuse layer: outputs of the last fuse layer are directly used as the outputs of the Target-Reference flow. 3) Transformer with complete select and fuse layers. All these models were trained on COCO-DIC and the results were shown in Table~\ref{table:abal_module}. 

\noindent \textbf{Results}. As can be observed in rows 2 and 3, two additional components can improve captioning performance consistently in terms of both accuracy and distinctiveness. Above all, our complete model achieves the most promising performance in all metrics.

\subsubsection{Influence of Group Similarity}
To quantify the influence of group similarity, we used different p when choosing K most similar images from $\mathcal{I}_f$, \eg, top-2 to top-6 \textbf{(top2-6)} group when setting p to 2. The results were reported in Table~\ref{table:abal_group}. 

\noindent \textbf{Results}. From Table~\ref{table:abal_group}(a), we can observe: Top1-5 group is surpassed by top3-7 group on DisCIDEr (101.4 v.s. 102.2), despite it having a marginal improvement on CIDEr (132.2 v.s. 132.0). The results indicate that the most similar reference group is not always helpful to group-level distinctiveness.

\begin{table*}[t]
\caption{Ablation study of group similarity on COCO-DIC and Flickr30K-DIC.}
\small
\subfloat[Comparison of different group trained with \textbf{TransDIC} on COCO-DIC]{
        \begin{tabular}{c|ccc}
        \hline
        \multicolumn{4}{c}{TransDIC at COCO-DIC}\\
        \hline
        Group  & B-4 & C & DisC \\[.1em]
        \hline
        top1-5  & 39.1 & \textbf{132.2} & 101.4\\
        top2-6  & \textbf{39.3} & 131.5 & 101.4\\
        top3-7  & \textbf{39.3} & 132.0 & \textbf{102.2} \\
        top4-8  & 38.7 & 130.7 & 101.5 \\
        \hline
        \end{tabular}}\hspace{1mm}
        \subfloat[Comparison of different group trained with GdisCap~on COCO-DIC]{
        \begin{tabular}{c|ccc}
        \hline
        \multicolumn{4}{c}{GdisCap at COCO-DIC}\\
        \hline
        Group  & B-4 & C & DisC \\[.1em]
        \hline
        top1-5  & \textbf{38.0} & \textbf{126.9} & 97.0\\
        top2-6  & 37.5 & 125.1 & 96.2\\
        top3-7  & 37.3 & 125.8 & 96.6 \\
        top4-8  & 37.6 & 125.3 & \textbf{97.1} \\
        \hline
        \end{tabular}}\hspace{1mm}
\subfloat[Comparison of different group with TransDIC on Flickr30K-DIC]{
        \begin{tabular}{c|ccc}
        \hline
        \multicolumn{4}{c}{TransDIC at Flickr30K-DIC}\\
        \hline
        Metric  & B-4 & C & DisC \\[.1em]
        \hline
        top1-5  & 29.4 & 64.5 & 40.6\\
        top2-6  & 28.8 & 62.4 & 39.3\\
        top3-7  & 30.1 & 65.1 & 41.6 \\
        top4-8  & \textbf{30.9} & \textbf{66.7} & \textbf{42.0}\\
        \hline
        \end{tabular}}\hspace{1mm}
\label{table:abal_group}

\end{table*}

\begin{table}[t]
    \caption{ Comparison of different masking strategies on COCO-DIC. }
    \vspace{-1em}
    \centering
    \begin{tabular}{{lcccccc}}
    \hline
    Model & B-1 & B-4 & M & R & C & DisC\\
    \hline    
    Baseline (TransDIC) & 81.6 & 39.3 & 29.2 & 58.5 & 132.0 & 102.2\\
    Zero Mask & 82.0 & 39.6 & 29.1 & 58.7 & 132.0 & 102.0 \\
    
    \hline    
        \multicolumn{7}{l}{\textbf{Instance-level Masking}} \\
        \cdashline{1-7}[1pt/1pt]
        Random Mask & 82.1 & 39.6 & 29.2 & 58.7 & 132.2 & 102.3 \\
        SimMask & 82.2  & 39.8 & 29.2 & 58.8 & 132.3 & 102.2  \\
        Grad-CAM & 82.3 & 39.8 & 29.2 & 58.8 & 132.3 & 102.3  \\

    \hline
        \multicolumn{7}{l}{\textbf{Image-level Masking}} \\
        \cdashline{1-7}[1pt/1pt]
        Random Image Mask & 82.2 & 39.7 & 29.1 & 58.8 & 131.5 & 101.4 \\
        Image Pool  & 82.0 & 39.7 & 29.1 & 58.5 & 131.5 & 101.4 \\
        Grad-CAM & \textbf{82.7} & \textbf{40.3} & \textbf{29.1} & \textbf{58.9} & \textbf{132.4} &\textbf{102.4}\\
    \hline
    \end{tabular}%

    \label{table:result_on_masking_coco}
\end{table}%

\subsubsection{Influence of Different Masking Strategies}
The experimental results of our proposed CL module on Table.~\ref{table:result_on_masking_coco}. Three kinds of baselines were designed: 1) ``Zero Mask'' strategy, \ie, all proposals in reference images will be masked, and the $\mathcal{X}_r^{mask}$ is set to zeros. 2) Instance-level random masking, each proposal in reference images has the 50\% to be masked. 3) Image-level random masking, each image will be masked according to 50\% probability. All above mentioned baselines and proposed masking strategies were trained on COCO-DIC.

\noindent\textbf{Results.} 1) As can be observed in row 1 and row 2, the proposed module can improve the performance of naive TransDIC, especially in BLEU metric (82.0 v.s. 81.6 at B-1 and 39.6 v.s. 39.3 at B-4). 2) Applying two kinds of masking strategies will further benefit the performance. Comparing the four kinds of masking strategies with their specially designed baselines, we observe steady improvements. Both \textbf{Grad-CAM} strategies applied on the two levels get the best performance, especially the image-level Grad-CAM. We see a large improvement at B-1 (82.7 v.s. 81.6) and B-4 (40.3 v.s. 39.3). The experiments demonstrate the promising results of the CL module.

\begin{table*}[t]
\caption{Ablation study of CL module on COCO-DIC.}
\vspace{-1em}
\subfloat[Ablation studies of DisReward.]{
    \begin{tabular}{ccc|cccc}
        \hline
        $\alpha_{b}$ & $\alpha_{c}$ & $\beta$  &  B-1 & B-4  & C & DisC\\
        \hline    
        0 & 0 & 0  & 81.6 & 39.3  & 132.0 & 102.2\\
        0.5 & 0 & 4  & 82.3 & 40.1  & 132.2 & 102.3\\
        0 & 1 & 4  & 81.5 & 39.0  & 133 & 102.9\\
        \hline
        0.25 & 0.5 & 4  & 82.1 & 39.6 & 132.2 & 102.3 \\
        0.25 & 0.5 & 2  & 82.3 & 39.6  & 131.6 & 102.2 \\
        0.25 & 0.5 & 8  & 82.4 & 39.7  & 132.4 & 102.3 \\
        \hline
    \end{tabular}}
\hspace{5mm}
    \subfloat[Ablation studies of masking thresholds.]{
        \begin{tabular}{cc|ccc}
        \hline
        Strategy & Threshold  & B-4  & C & DisC \\
        \hline
        \multirow{3}{*}{SimMask}  & 0.3 & 39.7 & 132.3 & 102.2\\
                                  & 0.5 & 39.8 & 132.3 & 102.2\\
                                  & 0.7 & 39.6 & 132.2 & 102.3 \\
        \hline
        \multirow{3}{*}{Grad-CAM} & 0.3 & 39.8 & 132.3 & 102.2\\
                                  & 0.5 & 39.8 & 132.3 & 102.3\\
                                  & 0.7 & 39.9 & 132.5 & 102.4 \\   
        \hline
    \end{tabular}}
\label{table:abla_contrastive}
\end{table*}

\subsubsection{Influence of DisReward Design}
In Sec.~\ref{sec:contrastive}, we design a mixed metric bleuder. In this section, we investigate the influence of these hyperparameters. We adopt the random masking strategy in all experiments and the results are shown in Table~\ref{table:abla_contrastive}(a).

\noindent \textbf{Results.} Compared with row 1, all experiment settings got performance gains. Specifically, we can find that: 1) When we deliberately emphasize a metric in \textbf{bleuder}, the final results will get the corresponding improvements in the metric. For example, when we set $\alpha_{b}$ to zero and $\alpha_c$ to 1, the model tends to focus more on CIDEr over the mixed-up one in row 4 (133 v.s. 132.2), however, this leads to the degradation on BLEU metric. Similarly, if we set $\alpha_c$ to zero, we can see a large gain on B-1 and B-4. So we finally use the mixed version of bleuder. 2) The margin parameter $\beta$ contributed less as the margin got larger. From 2 to 4, we saw a steady improvement, but turning 8 got a much less improvement.

\subsubsection{Influence of Masking Thresholds}

As mentioned in Sec.~\ref{sec:contrastive}, to remove ``useful information'' in reference images, when similarity or gradient score (similarity score and Grad-CAM in~\ref{sec:instance}) between proposals of target and reference images is higher than a threshold, the reference proposal will be masked to zero. In this section, we study the influence of threshold with on COCO-DIC, the experimental results are reported in Table~\ref{table:abla_contrastive}(b).

\noindent \textbf{Results.}
The higher the threshold is, the fewer proposals will be masked. Comparing row 2 and row 3, row 5 and row 6, we can observe that higher thresholds perform a little better. However, reducing the threshold from 0.5 to 0.3 did not get any improvements. We believe that it is because reducing the threshold at low volume (\ie, lower than 0.5) is useless, important proposals have already been masked previously (\eg, at threshold 0.7) because we sort the gradient scores.

\subsection{Qualitative Results}
\begin{figure*}[t]
\centering
\includegraphics[width=\linewidth]{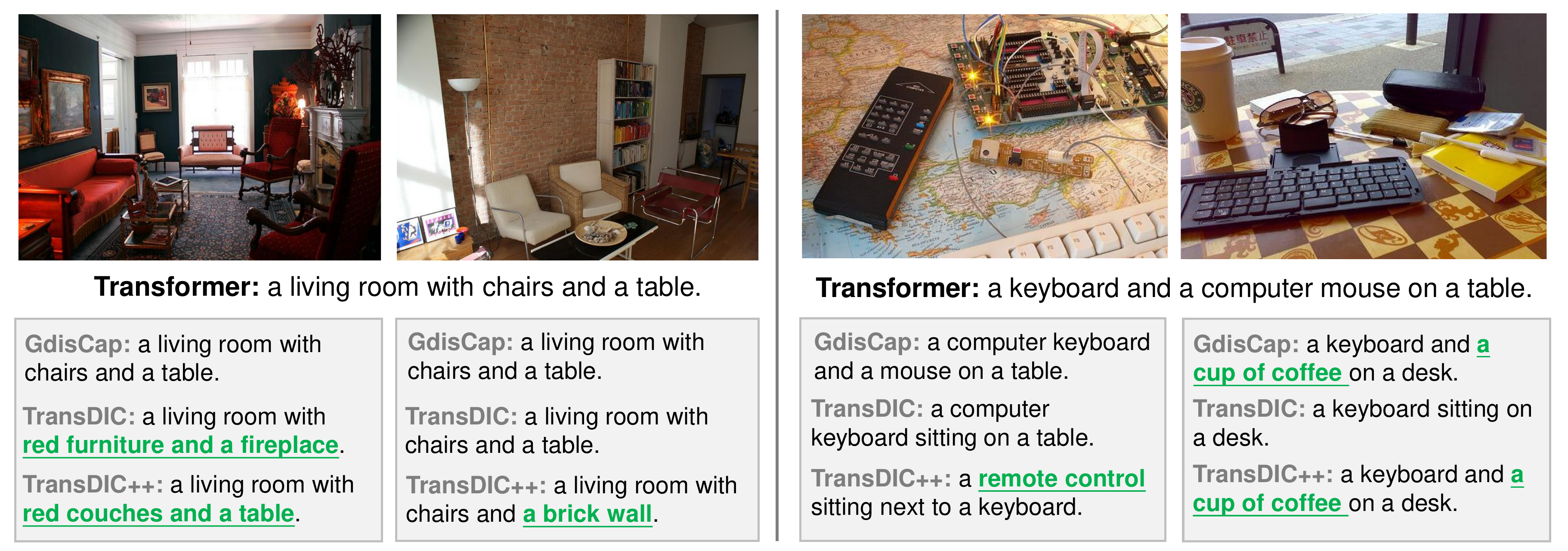}
\caption{Examples of generated captions for two similar images.  The \textcolor{mygreen}{green} words indicate the unique details in the images.}
\label{fig_6_example}
\end{figure*}

We illustrated the qualitative results of our proposed \textbf{TransDIC++} and compared it with the Transformer, Ref-DIC model GdisCap~\cite{wang2021group} and TransDIC\cite{mao2022rethinking} in Fig.~\ref{fig_6_example}. Naive captioning models generate identical captions for similar images. In contrast, our \textbf{TransDIC++} can describe the unique attributes and objects in the target image. For unique attributes, as shown in Fig.~\ref{fig_6_example}(left), \textbf{TransDIC++} precisely captures the unique attributes ``\texttt{red}'' of the couches to distinguish from the white couches in the similar image. Meanwhile, \textbf{TransDIC++} captures unique objects ``\texttt{brick wall}'' to tell apart from the left image. The results demonstrate that \textbf{TransDIC++} can generate distinctive captions in terms of unique objects and attributes.

\begin{figure*}[t]
\centering
\includegraphics[width=\linewidth]{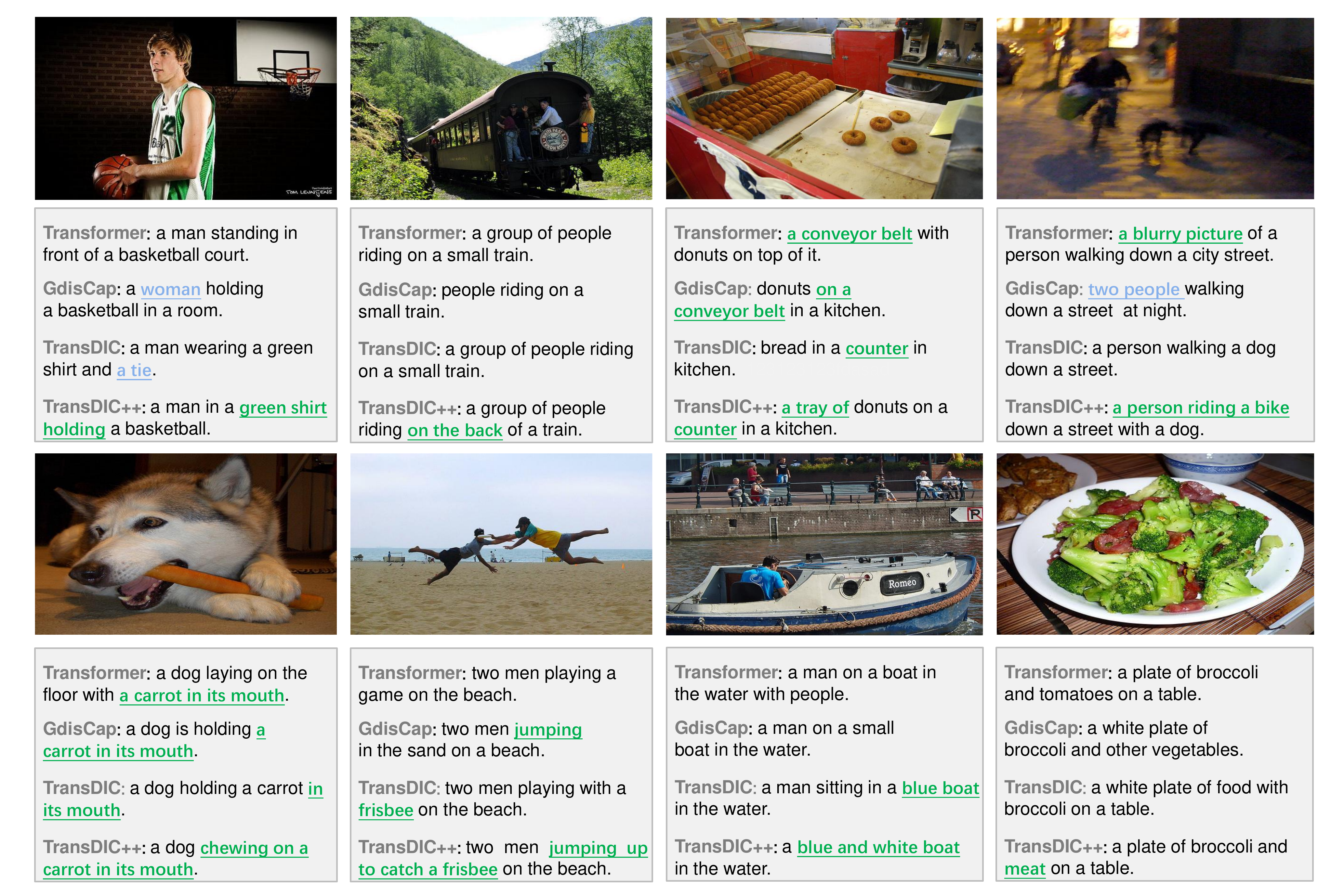}
\caption{Examples of generated captions for two similar images using Transformer, GdisCap, TransDIC, and \textbf{TransDIC++}. The \textcolor{mygreen}{green} words indicate the unique details in the images while \textcolor{myblue}{blue} denote the mistakes in the generated captions.}
\label{fig_s2}
\end{figure*}

More qualitative results are reported in Fig.~\ref{fig_s2}. In Fig.~\ref{fig_s2}, we compare our \textbf{TransDIC++} with Transformer, GdisCap and TransDIC. Remarkably, our model can precisely capture the unique objects and attributes in the image. For example in Fig.~\ref{fig_s2} (row1, column 1), GdisCap wrongly recognized the person as a woman and the TransDIC wrongly thought there exists a tie. The \textbf{TransDIC++} model precisely figured out the ``\texttt{green shirt}'' and the action ``\texttt{holding}''. In terms of numerals, in Fig.~\ref{fig_s2} (row2, column 4), \textbf{TransDIC++} not only predicted the right number ``\texttt{one}'' but also perceived the action ``\texttt{riding a bike}''. 

\begin{figure*}[t]
\centering
\includegraphics[width=\linewidth]{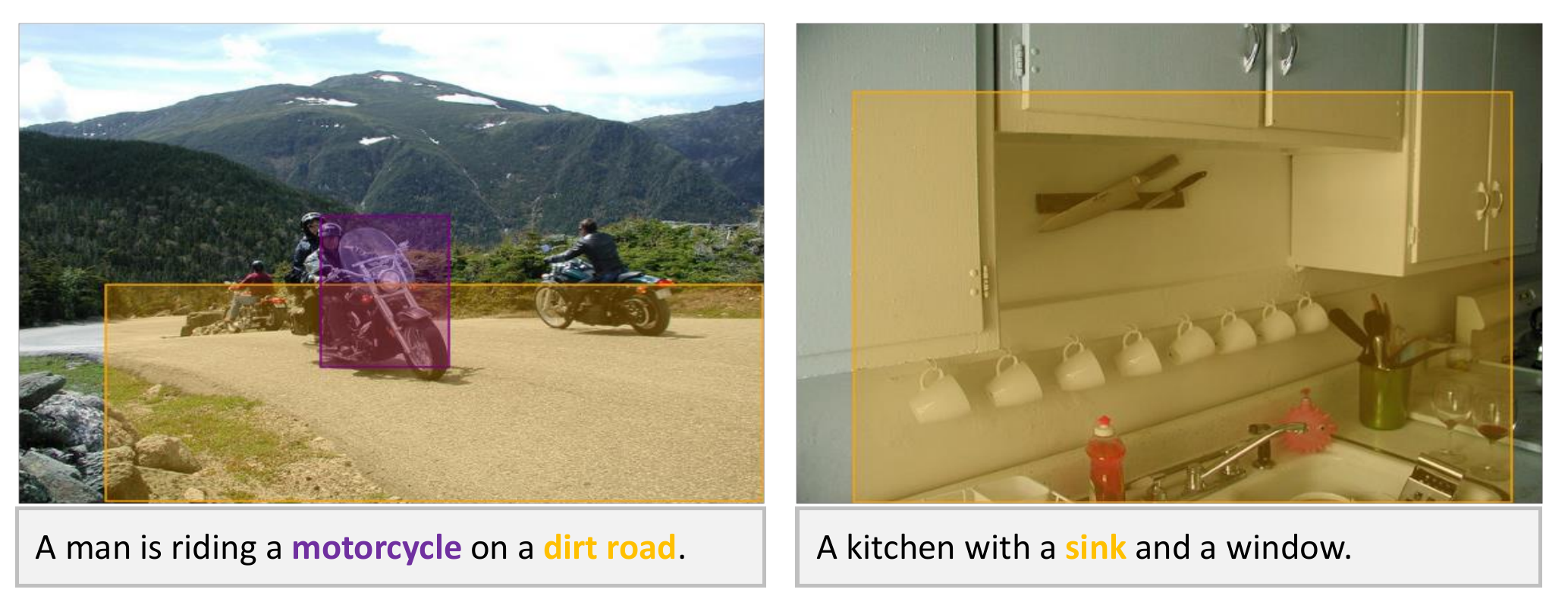}
\caption{Examples of Grad-CAM selected proposals in reference images.}
\label{fig_grad}
\end{figure*}

We also visualize the results of our Grad-CAM masking strategy in Fig.~\ref{fig_grad}. Leveraging a strong multi-modality model~\cite{chen2020uniter}, retrieved proposals in reference images align with the words that appear in the caption generated for the target image. As shown in the left of Fig.~\ref{fig_grad}, the multi-modality model will assign a higher grad score to the proposals of ``\texttt{motorcycle}'' and ``\texttt{dirt road}''.

\subsection{User Studies on DisCIDEr}
We conducted a user study to validate the effectiveness of \textbf{DisCIDEr} with 5 experts. We randomly selected 100 images (100 trials) from the test set. In each trial, an image and two captions (generated by Transformer and TransDIC, respectively) were displayed and these experts are asked to choose the better caption in terms of distinctiveness and accuracy. These captions which got more than 3 votes were regarded as human judgment. We calculated the agreements between human judgments and different metrics (\ie, whether humans and metrics give higher scores to the same caption). Results were reported in Table~\ref{table:user_study}. From Table~\ref{table:user_study}, we can observe that DisCIDEr achieves better agreement than both the accuracy-based metric CIDEr and the distinctiveness-based metric DisWordRate.

\begin{table}[h]
  \centering
    \caption{User study of different metrics}

    \begin{tabular}{c | ccc}
    \hline
    Metric & CIDEr~\cite{vedantam2015cider} & DisWordRate~\cite{wang2021group} & DisCIDEr  \\
    \hline
    Agreements & 58  & 62  & 64  \\
    \hline
    \end{tabular}%
    
    \label{table:user_study}
\end{table}%

\section{LIMITATIONS}
One possible limitation of our work is that if original human-annotated captions omit some objects or attributes, it will lead to: 1) our proposed two-stage matching mechanism may fail to collect object-/attribute- level similarity reference images. 2) our proposed DisCIDEr may degrade to existing CIDEr. We believe this omitting problem is due to the natural defect of datasets (COCO/Flickr30K): Since these datasets are annotated for general captioning tasks, human annotators may tend to simply describe the objects while ignoring their attributes (\eg, color) when there is no reference image.

\section{Conclusions}
In this paper, we comprehensively discussed two kinds of settings of the DIC task and demonstrate the important role of reference images. To this end, we proposed two new Ref-DIC benchmarks for this direction. Meanwhile, we developed a Transformer-based Ref-DIC baseline TransDIC. To further make full use of those reference images and investigate the influence of them, we designed a new contrastive learning based reward DisReward and developed two kinds of masking strategies. We conducted extensive experiments to verify the effectiveness of DisReward and the masking strategies. Meanwhile, we adopted our framework to other Ref-DIC models to demonstrate the flexibility of our proposed method. Moving forward, we are going to 1) extend our Ref-DIC into video domains, or 2) design stronger Ref-DIC models.

\bibliographystyle{ACM-Reference-Format}
\bibliography{tomm}

\end{document}